\pdfoutput=1

\documentclass[11pt]{article}

\usepackage[preprint]{acl}

\usepackage{times}
\usepackage{latexsym}
\usepackage{float}

\usepackage[T1]{fontenc}

\usepackage[utf8]{inputenc}

\usepackage{microtype}

\usepackage{inconsolata}

\usepackage{graphicx}

\title{FastPOS: Language-Agnostic Scalable POS Tagging Framework Low-Resource Use Case}

\author{Md Abdullah Al Kafi \\
  Daffodil International University \\Dhaka, Bangladesh \\
  \texttt{kafi.cse@diu.edu.bd} \\ \And
  Sumit Kumar Banshal \\
  Alliance University \\ Karnataka, India  \\
  \texttt{sumitbanshal06@gmail.com} \\}

\begin{document}
\maketitle
\begin{abstract}
This study proposes a language-agnostic transformer-based POS tagging framework designed for low-resource languages, with Bangla and Hindi serving as case studies. With only three lines of framework-related code, the framework was adapted to a new language, namely, from Bangla to Hindi. Displaying its effectiveness with minimal code change. Additionally, the framework achieves 96.85\% and 97\% token-level accuracy across POS categories, maintaining robust F1 scores despite dataset imbalance and linguistic overlaps in Bangla and Hindi, respectively. However, the performance discrepancy in a specific POS type highlights challenges in dataset curation. Moreover, the performance is due to the transformer used under the hood of this framework, which can be swapped with minimal code changes. The framework's modular, language-agnostic design and open-source design enable rapid adaptation with minimal code modification. By reducing model design and tuning overhead, researchers can prioritize linguistic preprocessing and dataset refinement, key tasks in advancing NLP for underrepresented languages.
\end{abstract}
\section{Introduction}
Parts of speech (POS) tagging is a core NLP task essential for higher-level applications such as parsing, sentiment analysis, machine translation, and many more. However, building an effective POS tagger for low-resource languages (LRLs) remains challenging due to the limited availability of annotated data, a lack of standardization, and high morphological complexity.

While language-specific models exist, they are hard to scale across diverse LRLs. This study introduces a novel language-agnostic, transformer-based POS tagging framework engineered explicitly for rapid adaptability with minimal code modification. Additionally, the framework supports modular integration and dataset flexibility across languages. We use Bangla and Hindi as a case study to demonstrate their effectiveness, achieving strong token-level accuracy despite class imbalance and linguistic overlaps, and also propose a standardized dataset for future investigations.
\section{Background} 
The severe lack of high-quality linguistic resources hinders the development of practical NLP tools. In LRLs, it is one of the primary issues. POS tagging tools are not too different. Significant progress has been achieved for high-resource languages, but for LRLs, it remains a formidable challenge. Data scarcity is prevalent and often limited to only thousands of tokens, a stark contrast to the millions available for high-resource languages \cite{McGiff2025}. This limitation significantly constrains the ability of neural models to learn robust generalizations.
Furthermore, many LRLs, including Bangla and Hindi, exhibit complex morphological structures, which leads to a high out-of-vocabulary problem \cite{Alam2017}. Additionally, the available datasets suffer from annotation inconsistencies, noise, and a lack of standardization \cite{Kim2015}. As a result, special processing steps require a considerable amount of time. Furthermore, finetuning of transformers usually offers strong performance in language-related workflows \cite{Devlin2019, Conneau, Liu2019}. Language-specific models, such as BanglaBERT, can capture language-specific features \cite{Bhattacharjee2022}. Moreover, tokinization plays an essential role in transformer performance due to the morphisms, and enriching supervision with sentence type helps models capture semantic nuance, benefiting tasks like question answering and dialogue systems \cite{Dang2024, majumdar}.
However, most existing transformer-based systems are not plug-and-play. They require substantial tokenization decisions and model engineering for each new language. This creates a barrier for scalable applications across LRLs. Researchers dedicate time and effort to maintaining the code base, where the primary focus should be on preprocessing, dataset standardization, and other related tasks. Recent benchmarking efforts (e.g., Indic-Transformers) confirm that transformer models, when properly fine-tuned, can achieve high F1 scores (e.g., 92.60 for Bengali) \cite{Sarker2021}. Yet, the ease of transferring these models across languages, as well as the use of modular frameworks that abstract underlying complexity, remains underexplored. 
To bridge this gap, a language-agnostic POS tagging framework that can adapt to new languages with minimal changes is essential. Such a system would support scalable deployment, rapid experimentation, and reduced engineering overhead, particularly critical in under-resourced environments. A modular design enables interchangeable transformer backbones, allowing researchers to experiment with different architectures (e.g., ByT5 for character-level modeling \cite{Dang2024}) without rewriting core logic. Minimal code changes ensure quick language adaptation, facilitating cross-lingual experimentation and comparative studies. This study proposes a language-agnostic, open-source POS tagging framework that enables high performance with minimal code modification. Such a system not only facilitates faster adaptation to new languages, evidenced by successful Bangla-to-Hindi transfer, but also shifts the focus towards linguistic preprocessing and dataset refinement, where most of the performance bottlenecks now reside.
\section{Methodology}
\begin{figure}[h]
    \centering
    \includegraphics[width=0.45\textwidth]{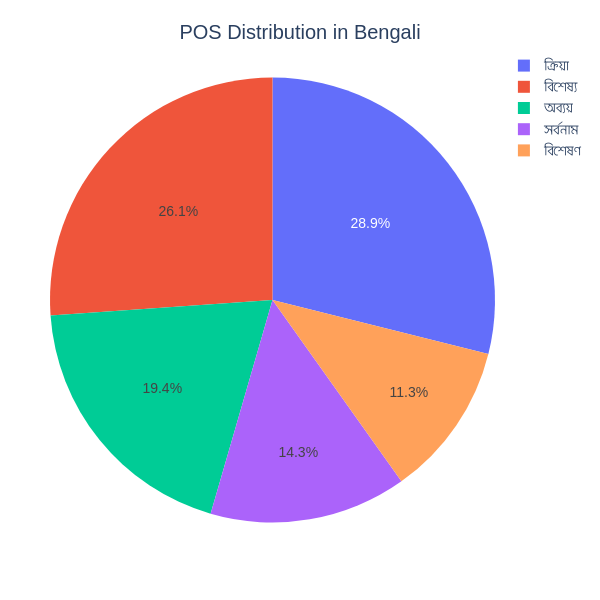} 
    \caption{Parts of Speech Distribution (Bangla)}
    \label{fig:posdist}
\end{figure}

\begin{figure}[h]
    \centering
    \includegraphics[width=0.45\textwidth]{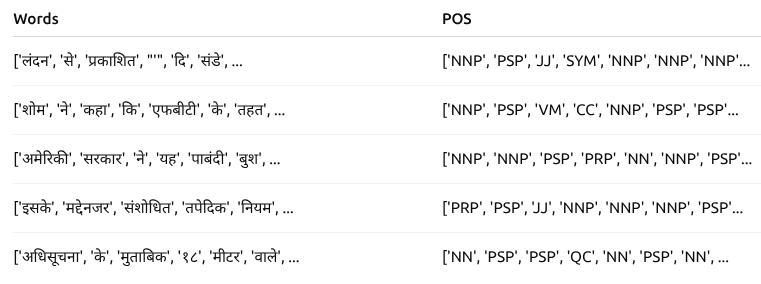} 
    \caption{Example of Modified Hindi Dataset}
    \label{fig:hindi}
\end{figure}

This study primarily utilizes a Bangla dataset (2696 instances across five main POS categories in Bangla) for Part-of-Speech (POS) tagging, complemented by a collected Hindi dataset (14963 instances across sixteen POS categories) to evaluate the framework's adaptability \cite{hinditask}. 
\begin{figure}[h]
    \centering
    \includegraphics[width=0.45\textwidth]{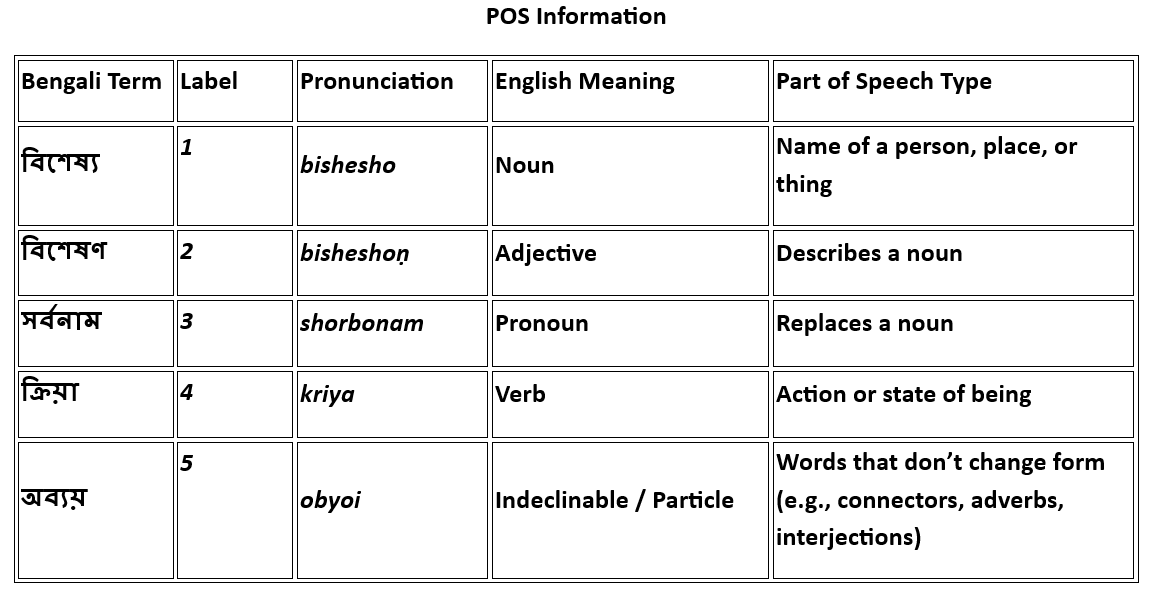} 
    \caption{Part of Speech Information (Bangla)}
    \label{fig:posinfo}
\end{figure}
Bangla dataset comprises 2,696 sentences (simple, complex, and compound), meticulously annotated with five primary POS types. Figure \ref{fig:posinfo} details these POS types, including English interpretations, which will be referred to as "POS type \#" hereafter. Figure \ref{fig:posdist} illustrates the imbalanced POS distribution inherent in the Bangla dataset, a common characteristic in real-world linguistic data, where POS types 1 and 4 collectively represent over half of the instances, similar to English.
Figure \ref{fig:hindi} illustrates an example of how this collected Hindi dataset was modified to align POS tags with words, fulfilling the framework's input requirements. The specific code used for this adaptability test is accessible on \cite{hinditask}
\subsection{Framework:} This POS tagging framework is designed for low-resource languages, leveraging the transformer model in its core. It supports custom datasets, training, prediction, and model management using PyTorch, HuggingFace, and Scikit-learn. Its modular and extensible architecture ensures easy integration into broader NLP pipelines with minimal overhead.
\subsubsection{UML Diagram:}
The following figure \ref{fig:umldiagram} illustrates the UML Diagram, which shows the architecture of the proposed POS tagging framework for low-resource languages, highlighting the interaction between core components that enable model training, prediction, and deployment.SentenceClassifier class details are not shared, as that is irrelevant to this study, and this framework is published as a class or module of the LowResNLTK framework \cite{anolow}.
\begin{figure}[h]
    \centering
    \includegraphics[width=0.45\textwidth]{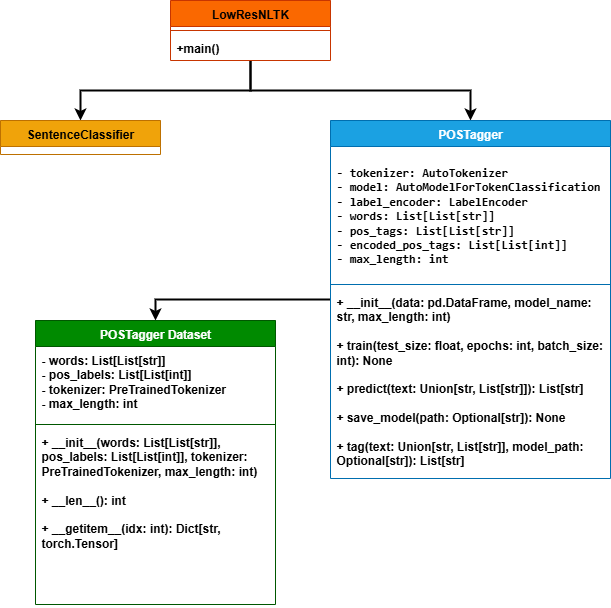} 
    \caption{UML Diagram}
    \label{fig:umldiagram}
\end{figure}
\subsubsection{Purpose:} 
\begin{itemize}
    \item Designed for Part-of-Speech (POS) tagging, especially for low-resource languages.
    \item Utilizes transformer-based models, giving it freedom to choose any HuggingFace transformer. 
\end{itemize}
\subsubsection{Key Features:} 
    \begin{itemize}
        \item Data Handling: Accepts data in Pandas DataFrame format with words and POS tags.
        \item Custom Dataset: Implements a PyTorch Dataset for tokenization and label alignment.
        \item Label Encoding: Uses sklearn’s LabelEncoder for mapping POS tags to numerical labels.
        \item Model Training: Supports training with HuggingFace’s Trainer and TrainingArguments.
        \item Prediction: Provides methods to predict POS tags for new text inputs.
        \item Model Saving/Loading: Can save and reload trained models and label encoders for reuse.
        \item Pretrained Model Support: Can load and use pretrained transformer models.
    \end{itemize}
    \subsubsection{Extensibility:}Easily adaptable to other transformer models and languages, and any number of POS tags.Modular design allows for integration into larger NLP pipelines.
    \subsubsection{Usages:}
        \begin{itemize}
            \item Train a POS tagger on a custom dataset easily.
            \item Predict POS tags for new sentences.
            \item Save and Load model for deployment.
        \end{itemize}
    \subsubsection{Integration:}Built on popular libraries like PyTorch, HuggingFace, scikit-learn, and pandas. Can be used as a standalone tool or easily integrated into other Python and NLP projects. It is open-source and easily modifiable to meet one's specific needs.
\section{Experiments}
This study used Bangla and Hindi as use cases to demonstrate the working principle. Here, we developed the system with the help of transformers, specifically "BanglaBERT" for Bangla and \textbf{Model name} with a token classification head, and an 80\%-20\% train-test split for each case study.

\begin{figure}[h]
    \centering
    \includegraphics[width=0.45\textwidth]{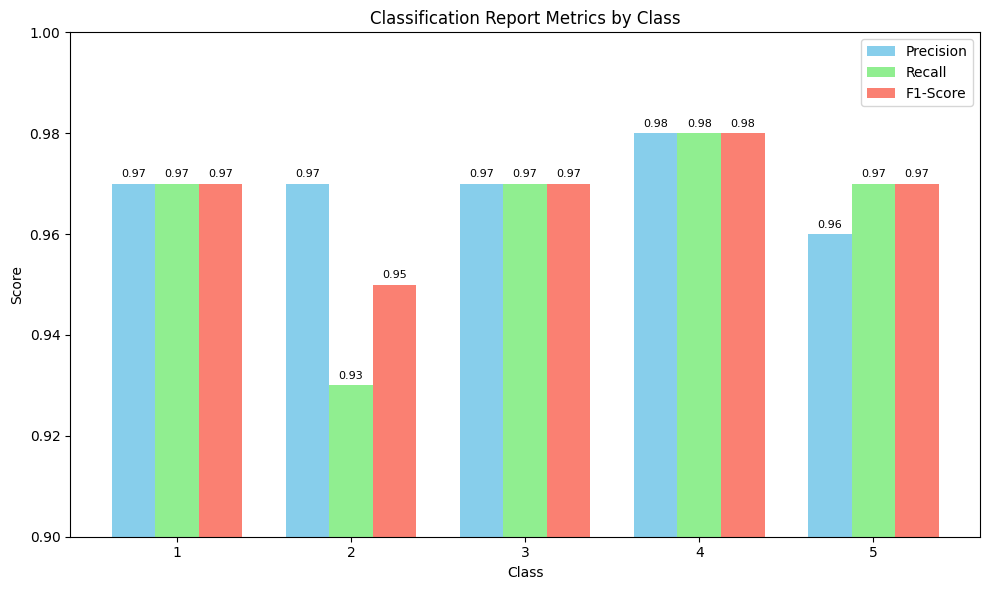} 
    \caption{Confusion Matrix Results Bangla}
    \label{fig:accuracy}
\end{figure}

\begin{table}[h]
\centering
\scriptsize
\setlength{\tabcolsep}{4pt} 
\caption{Classification report with token-level accuracy}
\begin{tabular}{lcccc}
\hline
\textbf{Class} & \textbf{Precision} & \textbf{Recall} & \textbf{F1-score} & \textbf{Support} \\
\hline
1 & 0.97 & 0.97 & 0.97 & 1265 \\
2 & 0.97 & 0.93 & 0.95 & 513 \\
3 & 0.97 & 0.97 & 0.97 & 622 \\
4 & 0.98 & 0.98 & 0.98 & 1376 \\
5 & 0.96 & 0.97 & 0.97 & 831 \\
\hline
\textbf{Accuracy}     &       &       & 96.85  & 4607 \\
\textbf{Macro Avg}    & 0.97  & 0.96  & 0.97           & 4607 \\
\textbf{Weighted Avg} & 0.97  & 0.97  & 0.97           & 4607 \\
\hline
\multicolumn{5}{l}{\textbf{Token-level accuracy:} 0.9685} \\
\end{tabular}
\label{tab:bangla}
\end{table}
Table \ref{tab:bangla} and Figure \ref{fig:accuracy} show that the token-level accuracy is 96.85\%, indicating reliable performance at the individual token level, which is particularly important in POS tagging. 
Focusing on the performance in detail reveals that the model performed well across all POS types, except for types two and five. For POS type two, low recall shows that the model is struggling to identify this type of POS.
\begin{table}[h]
\centering
\scriptsize
\setlength{\tabcolsep}{5pt}
\begin{tabular}{|l|c|c|c|c|}
\hline
\textbf{Category} & \textbf{Precision} & \textbf{Recall} & \textbf{F1-score} & \textbf{Support} \\
\hline
CC    & 0.71 & 0.50 & 0.59 & 10   \\
INTF  & 1.00 & 1.00 & 1.00 & 1    \\
JJ    & 0.25 & 0.23 & 0.24 & 26   \\
NEG   & 1.00 & 1.00 & 1.00 & 4    \\
NN    & 0.67 & 0.72 & 0.69 & 100  \\
NNP   & 0.71 & 0.63 & 0.67 & 38   \\
NST   & 0.00 & 0.00 & 0.00 & 1    \\
PRP   & 0.90 & 0.75 & 0.82 & 12   \\
PSP   & 0.93 & 0.85 & 0.89 & 47   \\
QC    & 1.00 & 0.40 & 0.57 & 5    \\
QF    & 1.00 & 1.00 & 1.00 & 1    \\
RB    & 0.50 & 1.00 & 0.67 & 1    \\
RP    & 0.33 & 1.00 & 0.50 & 1    \\
SYM   & 1.00 & 1.00 & 1.00 & 2666 \\
VAUX  & 0.83 & 0.88 & 0.86 & 34   \\
VM    & 0.73 & 0.83 & 0.78 & 46   \\
\hline
\textbf{Accuracy}      &       &       & 0.97 & 2993 \\
\textbf{Macro Avg}     & 0.72  & 0.74  & 0.70 & 2993 \\
\textbf{Weighted Avg}  & 0.97  & 0.97  & 0.97 & 2993 \\
\hline
\end{tabular}
\caption{Performance metrics (Precision, Recall, F1-Score, and Support) per category on the Hindi dataset.}
\label{tab:hindi}
\end{table}
In Table \ref{tab:hindi} and Figure \ref{fig:hindi}, despite achieving an impressive 97\% accuracy, the proposed framework, specifically the transformer, still faces the same issue as seen in Bangla.
\begin{figure}[h]
    \centering
    \includegraphics[width=0.45\textwidth]{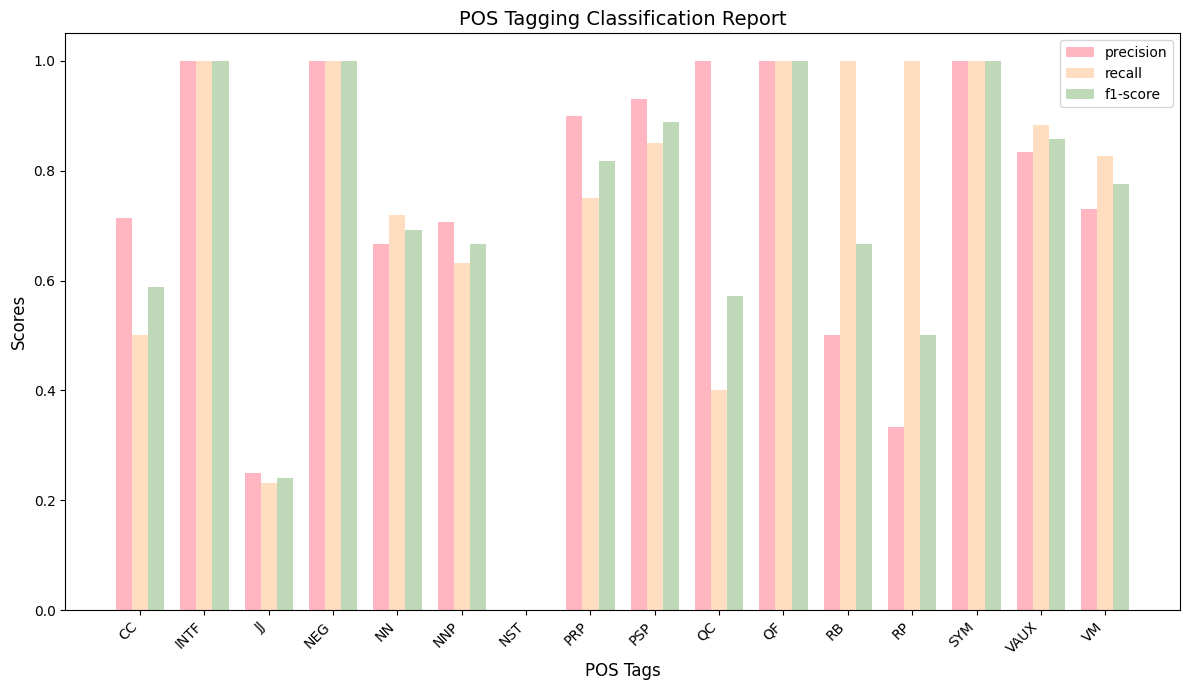} 
    \caption{Confusion Matrix Results Hindi}
    \label{fig:hindiconf}
\end{figure}
The performance analysis across both Bangla (Table \ref{tab:bangla}, Figure \ref{fig:accuracy}) and Hindi (Table \ref{tab:hindi}, Figure \ref{fig:hindiconf}) consistently reveals a critical limitation of the proposed transformer-based framework: its struggle with data sparsity and highly imbalanced classes. Despite achieving high overall accuracies (e.g., 96.85\% token-level accuracy in Bangla, 97\% overall accuracy in Hindi), the models exhibit apparent weaknesses in low-resource categories. For instance, Bangla's Class 2 exhibited lower recall due to fewer instances, while Hindi's 'NST' and 'RP' categories (with minimal support) yielded 0.00 F1-scores, and 'JJ' performed poorly. These discrepancies highlight that even advanced transformer models face significant hurdles with subtle grammatical distinctions and rare categories, often leading to under-prediction or false positives. This suggests that the performance limitations are not a flaw in the framework's design, but rather stem from the transformer's inherent inductive biases or insufficient pre-training on the specific linguistic nuances of such low-resource data.

\section{Conclusion}
This study presents a language-agnostic transformer-based POS tagging framework, using Bangla as a compelling case study. Data imbalance and Linguistic challenges, considering the model, yield an impressive accuracy of 96.85\% and 97\%, demonstrating a consistently high F1 Score across the five main POS categories. 
Performance discrepancies observed in POS types across both Bangla and Hindi highlight the limitations of current transformer architectures when confronted with the severe data sparsity and subtle linguistic overlaps prevalent in low-resource settings. With its language-agnostic architecture and minimal code dependency, the framework offers a scalable solution for morphologically rich, underrepresented languages in NLP.
\section{Limitations}
Despite its strong overall accuracy and adaptability for low-resource POS tagging, the proposed transformer-based framework has two key limitations. First, data imbalance causes under-predictions. Second, grammatical overlap with similar constructs leads to false positives. These challenges underscore the necessity for advanced data-centric approaches, such as multitask or few-shot learning, to address the inherent linguistic asymmetries and data sparsity in low-resource languages. We also plan to improve user adoption by aligning the framework's API with the scikit-learn interface.

\bibliography{custom}

@misc{anolow,
   title = {lowresnltk},
   url = {https://anonymous.4open.science/r/lowresnltk-7F6D/README.md}
}

@misc{hinditask,
   title = {Hindi Pos tagging test},
   url ={https://anonymous.4open.science/r/HindiPOSTaggigTest-752D/}
}

@article{Bhattacharjee2022,
   author = {Abhik Bhattacharjee and Tahmid Hasan and Wasi Uddin Ahmad and Kazi Samin and Md Saiful Islam and Anindya Iqbal and M. Sohel Rahman and Rifat Shahriyar},
   doi = {10.18653/V1/2022.FINDINGS-NAACL.98},
   isbn = {9781955917766},
   journal = {Findings of the Association for Computational Linguistics: NAACL 2022 - Findings},
   pages = {1318-1327},
   publisher = {Association for Computational Linguistics (ACL)},
   title = {BanglaBERT: Language Model Pretraining and Benchmarks for Low-Resource Language Understanding Evaluation in Bangla},
   url = {https://aclanthology.org/2022.findings-naacl.98/},
   year = {2022}
}

@article{Sarker2021,
   author = {Sagor Sarker},
   month = {1},
   title = {BNLP: Natural language processing toolkit for Bengali language},
   url = {https://arxiv.org/pdf/2102.00405},
   year = {2021}
}

@article{Dang2024,
   author = {Thao Anh Dang and Limor Raviv and Lukas Galke},
   month = {10},
   title = {Tokenization and Morphology in Multilingual Language Models: A Comparative Analysis of mT5 and ByT5},
   url = {https://arxiv.org/pdf/2410.11627v1},
   year = {2024}
}

@article{Liu2019,
   author = {Yinhan Liu and Myle Ott and Naman Goyal and Jingfei Du and Mandar Joshi and Danqi Chen and Omer Levy and Mike Lewis and Luke Zettlemoyer and Veselin Stoyanov and Paul G Allen},
   month = {7},
   title = {RoBERTa: A Robustly Optimized BERT Pretraining Approach},
   url = {https://arxiv.org/pdf/1907.11692},
   year = {2019}
}

@article{Devlin2019,
   author = {Jacob Devlin and Ming-Wei Chang and Kenton Lee and Kristina Toutanova Google and A I Language},
   city = {Stroudsburg, PA, USA},
   doi = {10.18653/V1/N19-1423},
   journal = {Proceedings of the 2019 Conference of the North},
   pages = {4171-4186},
   publisher = {Association for Computational Linguistics},
   title = {BERT: Pre-training of Deep Bidirectional Transformers for Language Understanding},
   url = {https://aclanthology.org/N19-1423/},
   year = {2019}
}

@inproceedings{majumdar,
    title = "{B}engali and {M}agahi {PUD} Treebank and Parser",
    author = "Majumdar, Pritha  and
      Alok, Deepak  and
      Bansal, Akanksha  and
      Ojha, Atul Kr.  and
      McCrae, John P.",
    editor = "Jha, Girish Nath  and
      L., Sobha  and
      Bali, Kalika  and
      Ojha, Atul Kr.",
    booktitle = "Proceedings of the WILDRE-6 Workshop within the 13th Language Resources and Evaluation Conference",
    month = jun,
    year = "2022",
    address = "Marseille, France",
    publisher = "European Language Resources Association",
    url = "https://aclanthology.org/2022.wildre-1.11/",
    pages = "60--67"
}

@inproceedings{conneau,
    title = "Unsupervised Cross-lingual Representation Learning at Scale",
    author = "Conneau, Alexis  and
      Khandelwal, Kartikay  and
      Goyal, Naman  and
      Chaudhary, Vishrav  and
      Wenzek, Guillaume  and
      Guzm{\'a}n, Francisco  and
      Grave, Edouard  and
      Ott, Myle  and
      Zettlemoyer, Luke  and
      Stoyanov, Veselin",
    editor = "Jurafsky, Dan  and
      Chai, Joyce  and
      Schluter, Natalie  and
      Tetreault, Joel",
    booktitle = "Proceedings of the 58th Annual Meeting of the Association for Computational Linguistics",
    month = jul,
    year = "2020",
    address = "Online",
    publisher = "Association for Computational Linguistics",
    url = "https://aclanthology.org/2020.acl-main.747/",
    doi = "10.18653/v1/2020.acl-main.747",
    pages = "8440--8451"
}

@article{Kim2015,
   author = {Young-Bum Kim and Benjamin Snyder and Ruhi Sarikaya},
   pages = {17-21},
   title = {Part-of-speech Taggers for Low-resource Languages using CCA Features},
   year = {2015}
}

@article{Alam2017,
   author = {Firoj Alam and Shammur Absar Chowdhury and Sheak Rashed Haider Noori},
   doi = {10.1109/ICCITECHN.2016.7860227},
   isbn = {9781509040896},
   journal = {19th International Conference on Computer and Information Technology, ICCIT 2016},
   month = {2},
   pages = {377-382},
   publisher = {Institute of Electrical and Electronics Engineers Inc.},
   title = {Bidirectional LSTMs - CRFs networks for bangla POS tagging},
   year = {2017}
}

@article{McGiff2025,
   author = {Josh McGiff and Nikola S. Nikolov},
   month = {5},
   title = {Overcoming Data Scarcity in Generative Language Modelling for Low-Resource Languages: A Systematic Review},
   url = {https://arxiv.org/pdf/2505.04531},
   year = {2025}
}

\end{document}